# Language guided machine action

**Feng Qi**[1,2]
**1.Alibaba-Group. 2. University of Oxford**
**qianlong.qf@alibaba-inc.com**

Robotic arm control based on reinforcement learn is similar to human habitual action based on basal ganglia, through which stimulus can elicit action automatically, but it cannot flexibly and consciously emit goal-directed action that is governed by human prefrontal cortex. Here, we build a hierarchical modular network called Language guided machine action (LGMA), whose modules process information mimicking human cortical network that allows to achieve multiple general tasks such as language guided action, intention decomposition and mental simulation before action execution etc. LGMA contains 3 main systems: (1) primary sensory system that multimodal sensory information of vision, language and sensorimotor. (2) association system involves Wernick and Broca modules to comprehend and synthesize language, BA14/40 module to translate between sensorimotor and language, midTemporal module to convert between language and vision, and superior parietal lobe to integrate attended visual object and arm state into cognitive map for future spatial actions. Pre- supplementary motor area (pre-SMA) can converts high level intention into sequential atomic actions, while SMA can integrate these atomic actions, current arm and attended object state into sensorimotor vector to apply corresponding torques on arm via pre-motor and primary motor of arm to achieve the intention. The high-level executive system contains PFC that does explicit inference and guide voluntary action based on language, while BG is the habitual action control center such as 'stop when traffic light is red'. We implemented LGMA model with autoencoder and LSTM modules in tensor-flow, and tested that it can perform various general tasks and achieve human-like motion control.

**Introduction**

To achieve human-like intelligent system, current AI pays attention to larger networks and training data sets. In the field of NLP, though state of the art models are continuously emerging, big model such as GPT3 is still a statistical model rather than an intelligent thinking machine [1, 2]. Human-like general language processing (HGLP) provides a hierarchical modular framework aiming to achieve strong artificial intelligence by simulating human brain architecture and information processing modes [3]. HGLP understands natural language via constructing mental scenario in visual autoencoder, instead of statistical modeling, which allows language to operate as script to manipulate imagined objects in the constructed mental scenario, such as achieving the machine thinking process guided by language [4]. This paper extends the HGLP framework to sensorimotor modalities, in order to let machine control its voluntary action consciously or verbally. Reinforcement learning is an effective means of motion control, simulating brain dopamine rewarding system underpinned by basal ganglia (BG), which can only perform habitual action subconsciously and require lots repeated practice beforehand [5]. On the contrary, PFC mediates voluntary action which is more flexible and can be governed by language consciously [6]. For example, BG let us stop at red light automatically, while PFC allow us to go through if safe. Here, we focus on the voluntary action mechanism of human brain and its implementation on machine.

The gap between machine and human action control is mainly reflected in the following aspects. (1) human with unified brain neural network can acquire and perform diverse tasks, such as writing, speaking and even coding etc. However, artificial neural network normally only expertise in one specific task or domain, such as alpha-Go [7] and BERT [8]. (2) As a human being, we can describe what we see and how we feel, can imagine the story portrayed in text and even can experience the feeling of actors. However, machine lacks efficient mechanism for integration of multimodalities of vision, language and sensorimotor. (3) Human being usually defines a series of actions as task or intention, and then plan and execute actions by thinking with such task or intention. For example, the intention of ‘fetch the ball’ can be decomposed into serial actions of ‘attend to the ball, then reach, hold, and finally pull it back’. But current motion control systems generate atomic action at one time. (4) People often do mental simulation before action execution, for example, we can hardly touch fire because of the anticipation of painful feeling. However, modern machine control systems directly take network output as real-time motor control signal, without extra components of result anticipation nor behavior modification. (5) To acquire a skill, machine needs repeated practice, especially failure experience. However, people have the capacity of forward verbal reasoning. So, after knowing the rule, we can play chess properly without any practice.

Solving these problems will bring machine closer to human-like intelligent action. In the HGLP model, we demonstrate how language is treated as script that can describe and control visual imagination rather than matching nor correlation statistics, and achieved 10+ different tasks based on the mnist scenario [9]. In this paper, we add the sensorimotor modules into HGLP according to anatomical structure of human brain, and achieve language guided machine action which allows robotic arms to perform diverse tasks, such as feeling pain, planning action, describing what have done, thinking with intention etc.

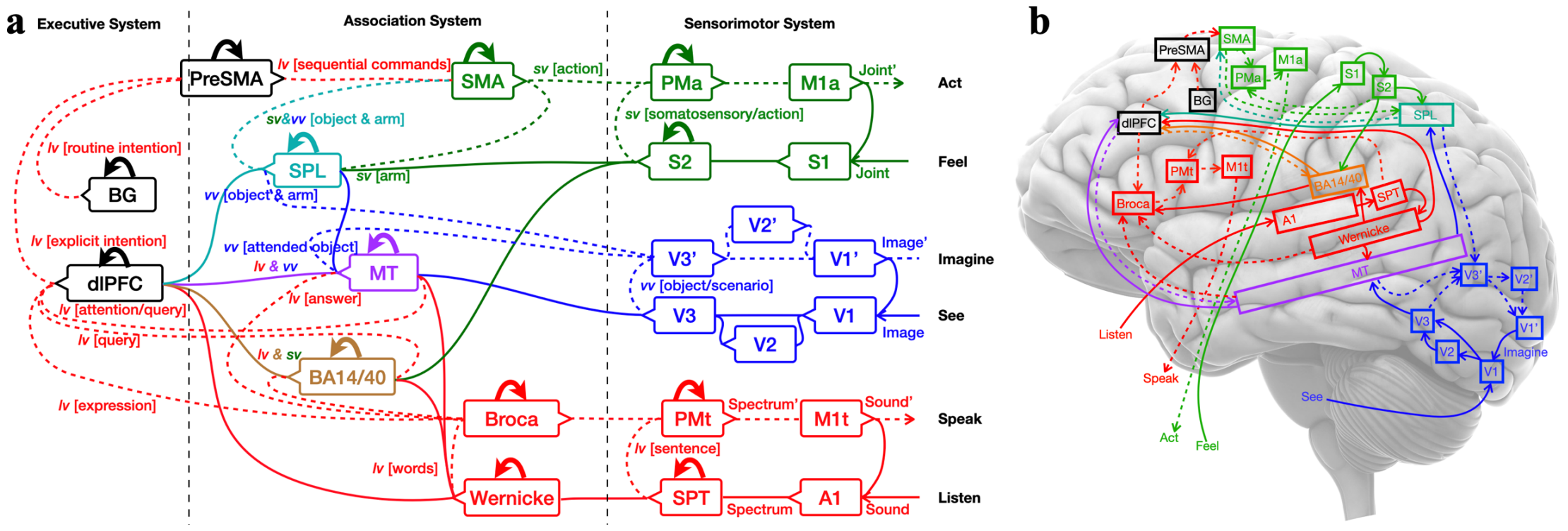

**Fig.1 | LGMA architecture. a,** LGMA is comprised of three hierarchical systems: sensorimotor, association, and executive systems. The low-level sensorimotor system has sensorimotor, visual and language subsystems that are made by autoencoders and trained with unsupervised learning in the early stage. The post-trained autoencoders could provide somatosensorimotor vector (*sv*) of gesture and pain, visual vector (*vv*) of images viewed, and language vector (*lv*) of sound heard, via its encoders. These *sv*, *lv* and *vv* are 256-byte vectors that can be reconstructed into joint torque, verbal articulation and visual imagination; Moreover, these vectors could be processed or be used as training data by higher level modules. In association system, there are Wernicke module that comprehends a sentence by decomposing it into phrases or words, Broca module that generates language expression with syntactical

rules, BA14/40 that names somatosensory state and motor actions, middle temporal (MT) module that functions as a translator between verbal and visual inputs, superior parietal lobe (SPL) that provides cognitive map of arms and attended object, pre-supplementary motor area (PreSMA) that decomposes intention into sequential commands to be executed by downstream modules, and supplementary motor area (SMA) that converts each command into arm action according to current arm and object states. The executive system consists of dorsal lateral prefrontal cortex (dlPFC) that can keep state vectors into its working memory, generate task response according to rules, and exert top-down control signal to lower level modules to properly interact with the environment, and Basal Ganglia (BG) that provides routine intention with respect to current situation. **b,** Anatomical connection of LGMA in human brain. Green represents somatosensorimotor path, red represents verbal paths, blue represents visual paths, purple represent visual and language association, orange is verbal and somatosensorimotor association, and cyan is visual and somatosensorimotor association. Solid represents feedforward paths, and dashed represents feedback paths. Each module has input and output ends, with curved arrow on top indicating LSTM, otherwise fully connected model.

**LGMA Architecture**

LGMA is hierarchical modular system, which follows Baddeley's model of working memory that consists of a central executive system used to control two slave systems (sensorimotor cortices): the phonological loop (PL) and the visuospatial sketchpad (VSS). Here, we extend Baddeley's model in two aspects: (1) add the association system with Wernicke and Broca modules to comprehend and generate language, BA14/40 to convert somatosensorimotor state into language, middle temporal (MT) to translate information between language and vision, superior parietal lobe (SPL) to integrate somatosensory state with attended objects, pre-supplementary motor area (PreSMA) to convert intention into serial atomic commands, and supplementary motor area (SMA) to generate action according to command and current arm-object state; (2) add the somatosensorimotor autoencoder consisting of sensory module S1 and S2, Premotor (PMa) and primary motor (M1a) of arm and hand, to let machine feel and behave. LGMA architecture is illustrated in Fig.1, which contains three hierarchies of sensorimotor, association, and executive systems. The low-level sensorimotor system is made of autoencoders and trained unsupervisedly to acquire abilities to process information of three modalities of somatosensorimotor, vision and language, respectively. The post-trained encoders could provide somatosensorimotor vector (*sv*) of action executed and pain felt, visual vector (*vv*) of images viewed or imagined, and language vector (*lv*) of sound heard or inner generated voice. The middle-level association system is implemented by long short-term memory (LSTM) models and trained supervisedly with salient events or facts represented by multi-modal vectors of *sv, vv* and *lv* to achieve diverse association functions across modalities. The high-level executive system consists of dorsal lateral prefrontal cortex (dlPFC) that can keep state vectors into its working memory, generate task response according to rules, and exert top-down control signal to lower level modules to properly interact with the environment, and basal ganglia (BG) module that provides routine intention with respect to current situation.

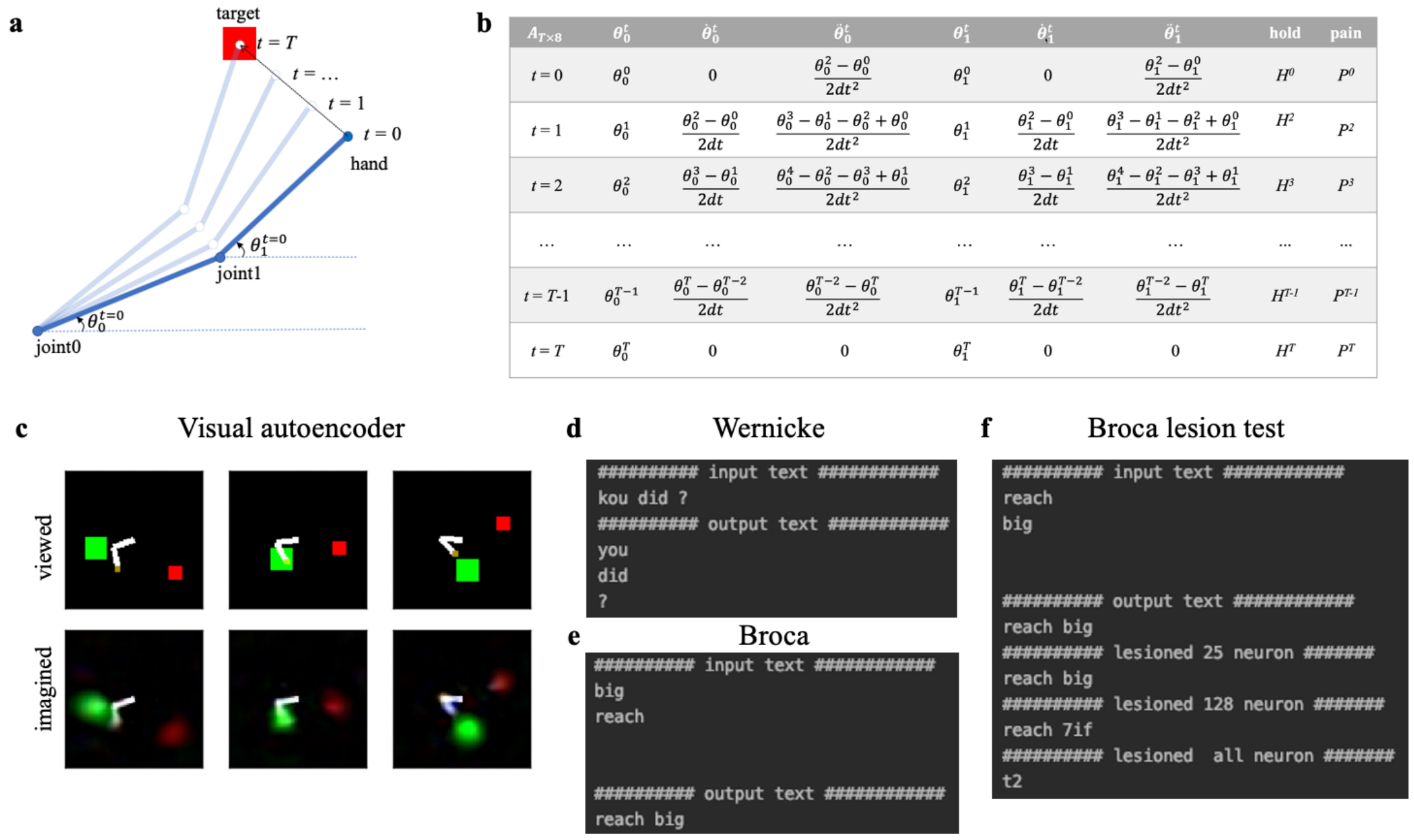

**Fig.2 | Somatosensorimotor, visual and language processing modules. a,** Straight reach action performed by LGMA arm can be decomposed into $T$ steps, with arm state explicitly expressed by two joint angles $\theta_i^t$, angular velocity $\dot{\theta}_i^t$, angular acceleration $\ddot{\theta}_i^t$, hand holding state $H^t$, and pain state $P^t$ at each time step $t$, where $i$ = 0 and 1 representing big and small arm, respectively. **b,** algorithm converts arm movement process into arm state matrix $\boldsymbol{A}_{T\times 8}$, which is used to train the somatosensorimotor autoencoder. **c,** visual encoder (V1V2V3 in Fig1a) converts the arm and objects viewed into 256-byte *vv*, which can be reconstructed into image imagined via visual decoder (V3'V2'V1'). **d,** Wernicke model comprehends sentence by decomposing it into phrase or word level *lv*, and correct wrong pronunciation. **e,** Broca can combine phrases or words into sentence *lv* syntactically for language generation. **f,** lesion of Broca neurons can cause Broca aphasia symptoms. With few (25 out of 256) neurons silenced, Broca can still generate complete sentence; the performance devastates but still readable as more neurons are lesioned; and could only articulate 't2' if all Broca neurons are silenced.

## Sensorimotor system

Sensorimotor system of LGMA is implemented and trained following HGLP. For visual processing, we construct a simple visual autoencoder (Fig. 1), with encoder part (V1V2V3) to extract a 256-byte visual vector (*vv*) from each image viewed or imagined, and a visual decoder part (V3'V2'V1') to reconstruct an image from a given *vv* (Fig. 2c). Here, we choose the three-layer structure with jumper between V1 and V3 rather than four-layer serial processing structure of HGLP, mainly because, in the robotic arm environment, this structure converges faster in training and its reconstructed image has better quality. For language processing, we construct a language autoencoder with encoder part (primary auditory cortex A1 and Sylvian parietal temporal (SPT)) to convert physical sound of a sentence into a 256-byte language vector (*lv*) and a language decoder part (premotor (PMt) and primary motor (M1t) of vocal cord, laryngeal and tongue areas) to articulate corresponding utterance from a given *lv*. The A1 module can convert physical sound into temporal spectrum signals, while M1a can articulate sound of the temporal spectrum signals (Methods). the SPT-PMt is implemented by a sequence-to-sequence module, where PMt aim to reconstruct accurately a temporal spectrum signal from *lv* encoded by SPT.

The somatosensorimotor system also plays essential role for human beings. For example, an insect falls on your neck, though you cannot see it, the somatosensory system can help you precisely locate the insect, perceive its size and even judge the risk level, and then the motor system of arms can repel the insect to avoid being bitten. The somatosensorimotor system of LGMA arms is similar to the language system. The language system can be viewed as sensorimotor system of vocal organs while the somatosensorimotor system can be used by deaf to deliver gesture language. LGMA somatosensorimotor system of arm is implemented by autoencoder with the encoder part (primary and secondary sensory cortex S1-S2) to convert the gesture and feeling of joints and skin into 256-byte somatosensorimotor vector (*sv*), and a decoder part (premotor (PMa) and primary motor (M1a) of arm and hand) to execute the limbic action from a given *sv*. Arm action such as reach red block in Fig.2a-b is a temporal process, that can be divided into $T$ time steps, and the arm state at each step can be described by an 8-dimensional vector $[\theta_0^t, \dot{\theta}_0^t, \ddot{\theta}_0^t, \theta_1^t, \dot{\theta}_1^t, \ddot{\theta}_1^t, hold^t, pain^t]$ representing joint angle, angular velocity, angular acceleration, hold stat, and pain value at step $t$ respectively. For straight reach action, given the initial and target hand position, Fig.2b can be used calculate all arm state parameters of the whole action process in term of arm state matrix $\boldsymbol{A}_{T\times 8}$. Note only the *acceleration* and *hold* are used as actuator signals to control two arms and hand, and we suppose the initial and final arm velocity are kept still. The arm state matrix $\boldsymbol{A}_{T\times 8}$ is used to train the somatosensorymotor autoencoder unsupervisedly to acquire the ability to perceive arm-hand state and execute actions.

**Association System**

Human association cortices consist of extensive territories of gray matter located between sensorimotor and executive cortices to integrate multimodal sensorimotor input and facilitate various tasks execution. In the LGMA network, we primarily consider those modules and functions related to arm actions. Wernicke and Broca modules are implemented similarly to HGLP, but in LGMA the enhanced 256-byte *lv* is used for information transmission and a vocabulary involving more action related words is used for training these modules. Wernicke is responsible for language comprehension by decomposing sentence into word or phrase level *lv* and correct the wrong pronunciation such as from 'kou' to 'you' in Fig. 2d. Broca is responsible for synthesizing sentence *lv* by syntactically rearranging words or phrases from upstream modules (Fig. 2e). So, lesion to Broca would cause problems in language production. As is displayed in Fig. 2f, when a small number (25 out of 256) of Broca neurons are silenced (activation set to zero), LGMA can still articulate complete sentence with the lesioned *lv*, due to the population coding mechanism; when a larger proportion neurons (128 out of 256 neurons) are lesioned, the articulation system (PMt-M1t) would generate unreadable utterance; When most of Broca neurons (256 out of 256) are lesioned, LGMA demonstrates remarkable symptom of Broca aphasia (i.e. output 't2' no matter what input contents are given by upstream modules) quite similar to patient 'Tan'.

Human MT takes up a large area (BA 21, 37, 38, 39) between the language-related superior temporal cortex and visual-related inferior temporal cortex. Its anterior part processes the objectness-related information, such as naming the color, shape or identity of a visual object, or imagine the story or object regarding to language heard. So, lesion to these parts will cause semantic dementia. The posterior part of MT processes spatial related information, such as naming the position or orientation of an visual object, or imagine objects arrangement or motion according to language heard. So, lesion to these parts will cause Alzheimer disease. HGLP paper demonstrates how vision and language are interacted based on MT module and mnist environment for object recognition, imagination and language guided attention, etc. In reality, every moment, thousands of visual objects flow into our visual system, however, we could only handle few of them highlighted by attention mechanism. In LGMA, MT is trained to have the voluntary attention mechanism guided by consciousness or language, such as outputting features of small square when hearing *lv* 'small' (Fig. 3a). This is different from reflex attention such as attention induced by sudden shining object, nor routine attention such as unconsciously watching traffic light before driving across road. MT can not only recognize object from viewed scenario according to verbal size, color and identity (square or arm), but also perceive and predict object motion. For example, *lv* 'initial' let MT output the initial position of the attended object, 'predict' let MT output the future object position. MT has two modalities output of *vv* and *lv* of attended object, that can be either future processed by higher level modules, or imagined and articulated by visual and verbal decoders, respectively.

Human SPL receives a great deal of visual input as well as somatosensory input from arm and hand (Fig.1), which allow SPL to function as egocentric cognitive map to judge where is surrounding object relative to the body. Fig.3b demonstrates that LGMA visual system viewed big, small squares and its arms, however, MT with *lv* attention mechanism only allowed the 'big' object to pass to SPL. Meanwhile, SPL also could access the arms state information *sv* from S2, so the output signal of SPL contained state information of both its own arms and attended object, which could be visually decoded via V3'V2'V1' as imagined cognitive map in fig.3b bottom row. This is especially important for spatial object reaching etc. tasks. The functionality of module SPL could explain many neurological phenomena, such as lesion of SPL would cause neglection of objects or body parts in the contra-lateral space. This module is implemented with LSTM and trained with arm *sv* and attended object *vv* as input and combined *vv* as target output.

The pre-supplementary motor area (preSMA) module links between executive and association systems, and functions as a converter from high hierarchical intention to executable atomic actions. The high-level intention signals can either be explicit inferred from dorsal lateral prefrontal (dlPFC), such as 'fetch bread' when hunger, or be habituated intention coming from basal ganglia (BG), such as 'stop' when traffic light turns red. PreSMA is responsible to convert these intentions in term of *lv* into sequences of atomic actions for future execution. For example, in case of 'fetch bread', pre-SMA would decode *lv*[fetch] into *lv*[reach] *lv*[hold] *lv*[pull] *lv*[fetch], and could convey high level *lv*[stop] in middle of sequential action execution. SMA locating posterior to preSMA, receives sequence of atomic commands from preSMA and cognitive map information (i.e. states of arm and attended object) from SPL, and computes corresponding *sv* signals to govern arms and hand actions. Fig.3d demonstrates that when attention signal of 'green' is given, there is no torque response to the object, but SPL implicitly represent the attention information in the output signal of *vv* [attended object & arms]. However, when SMA receives *lv* [reach], it converts the *lv* command into action *sv* according to current arms and object states. This action *sv* can be translated into joint 4 time-step torques via PreMa-M1a to let arm reach the object.

BA14/40 is a multimodal integration module between *lv* and *sv*. It can translate the painfulness from skin sensor via *sv* into *lv* expression of ‘very painful’ or ‘no pain’ which can be articulated through verbal decoder PMt-M1t. Since the *sv* represents a whole action process (i.e. multiple time-steps of actions), BA14/40 can convert behavior process just performed into language expression, such as answer question of ‘what you did ?’ or ‘you did ?’ as is shown in Fig. 3f. BA14/40’s output *lv* can be articulated as pull or push etc. via PMt-M1t according to if the yellow square is moved towards or outwards the center (‘self’). As in Fig.3g, the BA14/40 can also identify action sequence such as fetch = reach + hold + pull + release etc.

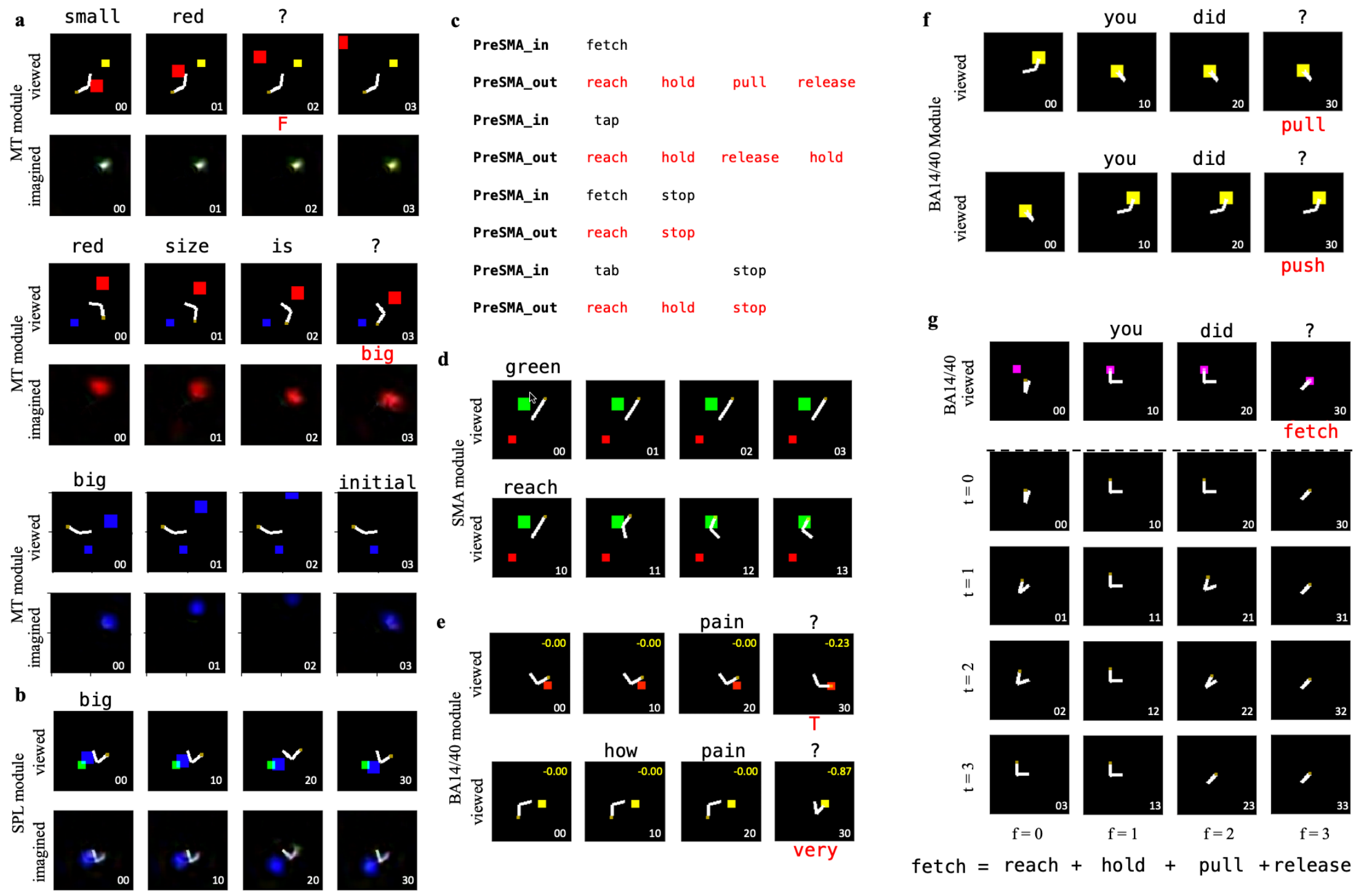

**Fig. 3 | functions of association modules**. **a,** middle temporal (MT) module attends to objects highlighted by language. The *lv* ‘big’ makes MT focusing only on the big blue square, while ‘initiate’ let MT remember where the big come from. **b,** Superior temporal lobe (SPL) module combines object signal from MT and arm state signal from S2 into its cognitive map in term of *vv*, that can be reconstructed via visual decoder. **c,** Pre-supplementary motor area (pre-SMA) receives either explicit intention from PFC or routine intention from basal ganglia (BG) in term of *lv*, and convert the intention into sequential atomic actions for future execution. Such as *lv* ‘fetch’ is converted into ‘reach > hold > pull >release’, and PreSMA can convey high level ‘stop’ command in middle of sequential action execution. **d,** Supplementary motor area (SMA) module receives atomic action from PreSMA, cognitive map of arm and attended object from SPL, and generate *sv* to control the arm action via PreMa-M1a. The top panel (frame 0) shows that SMA does not provide torque response to the ‘green’ object, while bottom panel

(frame 1) shows that proper torque is generated by SMA to let arm ‘reach’ the target object. **e,** BA14/40 is the translator between *sv* and *lv*. It can describe whether hand is pain and the painfulness according to the *sv* converted from sensory state $P^t$. **f.** BA14/40 can also answer query of ‘you did ?’ according to the action performed, such as reach, pull, push, hold, release, retract, etc. **g,** BA14/40 can also identify action sequence such as fetch = reach + hold + pull + release. Note, each frame is a series of actions that are deployed by 4 time steps below the dashed lines.

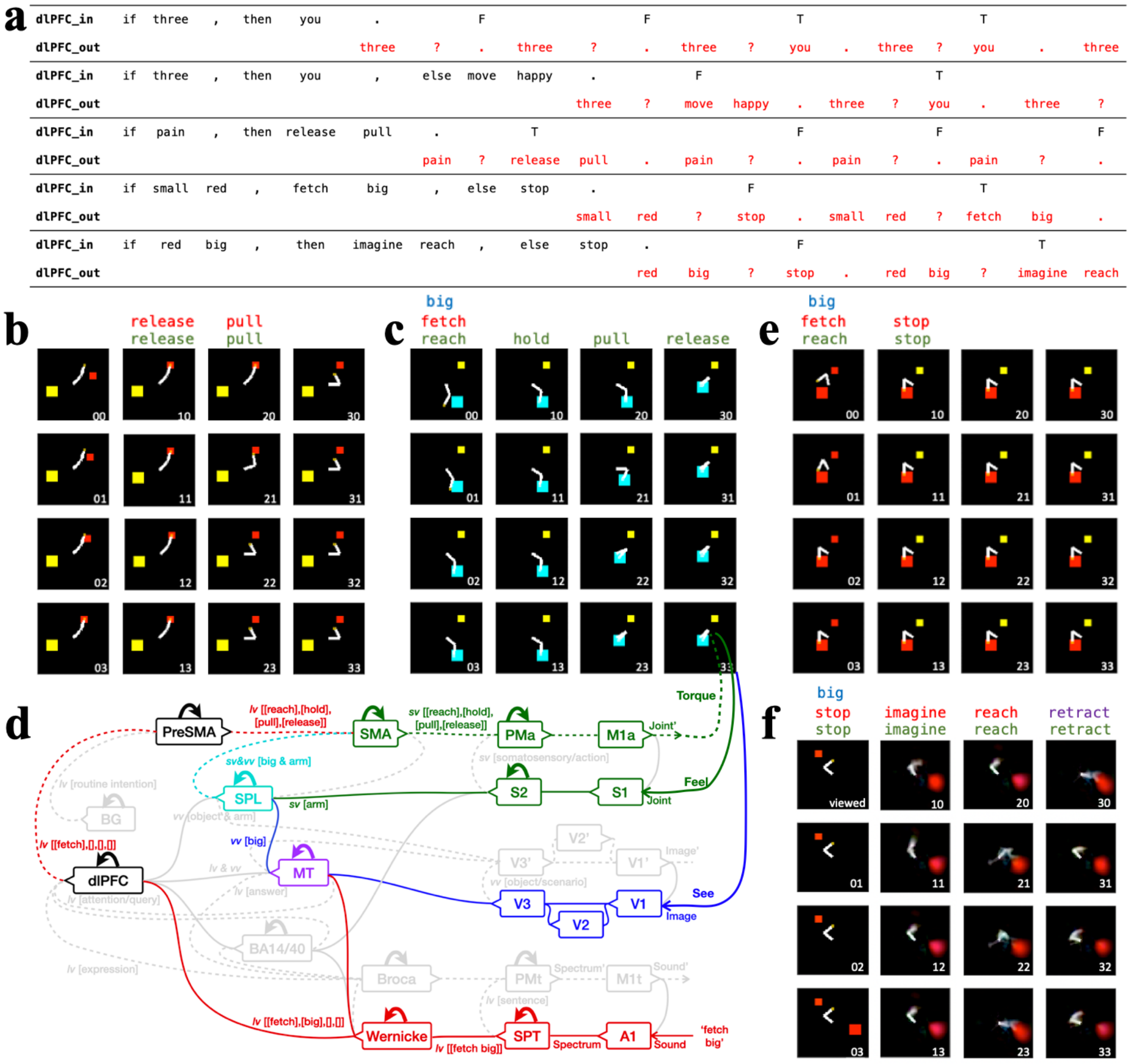

**Fig. 4 | Language guided voluntary action**. **a,** dlPFC module consciously selects task based on current condition. dlPFC is trained to perform ‘if *condition*, then *intention A*, else *intention B*’, where condition, *action A* and *B* are random words selected from vocabulary. After hearing ‘If then’ sentence, dlPFC will check whether current system meets ‘condition ?’, and association modules will answer the queries with T indicating ‘True’, while F indicating ‘False’, based on which dlPFC will execute *intention A* or *B*. **b,** pain reaction task, after hearing Fig. 4.row3, dlPFC will repeatedly check ‘pain ?’ with BA14/40. When red square representing hot object touch bot hand, the pain feeling triggers ‘release pull’ intention, which is converted into action sv via PreSMA-SMA and finally executed as joint torque via PMa-M1a. Note, red word is intention *lv* from dlPFC or BG, green are atomic intention *lv* from pre-SMA, blue is attention

*lv* to MT module, and purple is intention *lv* from BG. **c,** Fetch intention is decomposed by Pre-SMA into sequential actions (reach, hold, pull, release), and executed via PMa-M1a. **d,** Urgent stop intention. dlPFC (Fig. 4a row4) generates 'fetch big' intention in 'small red' condition, while sends 'stop' to terminate the sequential actions immediately after small square turns yellow. **e,** Imagery simulation before action execution. dlPFC (Fig. 4a row5) generates 'IMAGINE reach' intention after 'green big' square occur, the 'reach' *lv* can be fed into SPL with object *vv* to form the cognitive map *vv* of hand and attended object, which can be reconstructed at retina layer via visual decoder V3'-V2'-V1'. The imagined green touching event may be a disgusting scenario, based on which BG generates 'retract' intention to let arm go back to original postion. Note, since we didn't implement BG yet, the last step BG intention is artificially generated.

**Executive system**

**dlPFC module.** The executive functions of the human brain are located in prefrontal area, which is considered to be orchestration of thoughts and actions to achieve internal goals. For simplicity, LGMA merely involves a dlPFC module with executive related functionalities including, but not limited to, task/rule recognition, attention, working memory, query, and inference. Human dlPFC is the neural substrate for task and rule identification and representation. Conditional selection 'if then' is the abstract expression of rule-based reasoning, which allows people to flexibly handle numerous tasks in real-time, such as syllogism reasoning. That is why all programming languages have 'if then' statement. We trained an LSTM to acquire such conditional selection (or abstract reasoning) capacity, with training data generated by template 'if condition, then action A, else action B', where condition, action A and action B were assigned with one or two random words (Methods). As displayed in Fig. 4a, after dlPFC hearing the statement, it started to output query 'condition ?'; if received the answer 'T', dlPFC output action A, otherwise action B. After sufficient training, conditional selection in dlPFC could also be generalized to words beyond the training vocabulary. Fig.4b demonstrates a pain reaction task. After hearing 'if pain, then release pull' (Fig. 4a.row3), dlPFC will repeatedly check 'pain ?' through BA14/40. Here, the red square represents hot object. When robot hand touch the hot object, skin sensor will acquire $pain^t > 0$ which will trigger pain feeling expression in BA14/40, and then 'release pull' intention in dlPFC, which will be converted into action *sv* via PreSMA-SMA, and finally executed as joint torque via PMa-M1a. When people hunger, dlPFC will trigger fetch object intention. Fig.4c simulates the situation: dlPFC generate *lv* [fetch] intention, and preSMA converts it into sequential atomic actions of reach, hold, pull and release that could be executed through PMa-M1a. Fig.4d reveals the streaming process during fetch big task. The vision encoder system (V1V2V3) converts the visual situation into *vv*, A1-SPT converts the sentence 'fetch big' into 256B *lv*, and somatosensory S1-S2 converts arms states into *sv*. *lv*[big] can highlight the cyan square in MT while filtered out other visual objects including arms. SPL integrates both *vv*[big] from MT and *sv*[arm] from S2 and computes the cognitive map of *vv*[attended object & arms]. dlPFC follows the utterance 'fetch' and explicitly generates the *lv*[[fetch][][][]] intention to preSMA, which then convert it into *lv*[[reach][hold][pull][release]] for downstream execution. SMA takes the atomic commands from preSMA and cognitive map from SPL, and outputs corresponding action *sv*[[reach][hold][pull][release]]] which can be converted into joints torques to achieve the final 'fetch big' task. Fig4.f shows the imagery simulation before action execution. dlPFC (Fig. 4a row5) generates 'IMAGINE reach' intention after 'red big' square occur, the 'reach' *lv* can be fed into SPL with object *vv* to form the cognitive map *vv* of hand and attended object, which can be reconstructed at retina layer V1 via visual decoder V3'-V2'-V1'. The imagined red touching event is painful in experience, based on which BG generates 'retract' intention to let arm go back to original position before being hurt.

## Conclusion

The LGMA architecture can many unique traits: (1) like in HGLP, LGMA understands external language by constructing corresponding an imagery scenarios, it can also use language as script to manipulate the mental scenarios, and articulate the mental scenario via PMt-M1t. (2) LGMA provides a uniform architecture to integrate multimodalities of sensorimotor, vision and language together inspired by human brain system. It can describe the painful feeling and what action has been performed etc. (3) In contrast to previous Reinforcement learning that uses reward to guide robot action, LGMA use language to guide machine action. So even though naiive LGMA don't see the value of one action, it can still perform certain tasks instructed by external language. (4) LGMA can use intention to guide sequential actions. for example when generation of intention of 'fetch banana' it already prepared all 'reach hold pull and release' steps, and each action real-time depends on its current cognitive map of attended banana and arms states. (5) LGMA can imagine the future results of actions before executing, so it can stop actions before harmful results come. Just like we are afraid of bungee jump even we know it is safe.